\documentclass[10pt]{article} % For LaTeX2e
\usepackage[preprint]{tmlr}

\usepackage{amsthm}
\usepackage{times}
\usepackage{soul}
\usepackage{url}
\usepackage[utf8]{inputenc}
\usepackage[small]{caption}
\usepackage{graphicx}
\usepackage{hyperref}
\usepackage{amsmath}
\usepackage{paralist}
\usepackage{amssymb}
\usepackage{booktabs}
\usepackage{algorithm}
\usepackage{subcaption}
\usepackage{algorithmic}
\usepackage[inline]{enumitem}
\usepackage[switch]{lineno}
\usepackage{natbib}

\newtheorem{theorem}{Theorem}
\newtheorem{lemma}[theorem]{Lemma}
\newtheorem{corollary}[theorem]{Corollary}

\renewcommand{\cite}{\citep}

\title{Solving Truly Massive Budgeted Monotonic POMDPs with Oracle-Guided Meta-Reinforcement Learning}

\author{\name Manav Vora \email mkvora2@illinois.edu \\
       \addr Department of Aerospace Engineering\\
       University of Illinois Urbana Champaign\\
       Urbana, IL 61801, USA
       \AND
       \name Jonas Liang \email junhang2@illinois.edu \\
       \addr Department of Mathematics\\
       University of Illinois Urbana Champaign\\
       Urbana, IL 61801, USA
       \AND
       \name Michael N. Grussing \email Michael.N.Grussing@erdc.dren.mil \\
       \addr  Engineer Research and Development
        Center\\
        U.S. Army Corps of Engineers\\
       Champaign, IL 61822, USA
       \AND
       \name Melkior Ornik \email
       mornik@illinois.edu \\
       \addr Department of Aerospace Engineering\\
       University of Illinois Urbana Champaign\\
       Urbana, IL 61801, USA}

\begin{document}

\maketitle

\begin{abstract}
Many real-world decision problems, ranging from asset-maintenance scheduling to portfolio rebalancing, can be naturally modelled as budget-constrained multi-component monotonic Partially Observable Markov Decision Processes (POMDPs): each component’s latent state degrades stochastically until an expensive restorative action is taken, while all assets share a fixed intervention budget.  
For a large numbers of assets, deriving an optimal policy for this joint POMDP is computationally intractable. To tackle this challenge, we prove that the value function of the associated belief-MDP is \emph{budget-concave}, which allows an efficient two-step approach to finding a near-optimal policy. First, we approximate the optimal cross-component budget split via a random-forest surrogate of each single-component value function. Second, we solve each resulting budget-constrained single-component POMDP with an oracle-guided meta-trained Proximal Policy Optimization (PPO) policy: value-iteration on the fully observable counterpart yields an oracle that shapes the PPO update and greatly accelerates learning. We validate our method through experiments in two disparate domains: (i) preventive maintenance for a large-scale building infrastructure containing 1,000 components, and (ii) portfolio risk management under debit-only loss-budget constraints, where each asset’s latent budget depletes with market losses and can only be replenished through costly recapitalization. Results show that our method consistently achieves longer component survival times and enhanced portfolio viability than both baseline heuristics and vanilla PPO. Furthermore, our approach maintains linear scalability in solution time with respect to the number of components.

\end{abstract}

\section{Introduction}\label{sec:intro}

Partially Observable Markov Decision Processes (POMDPs) offer a principled framework for sequential decision making under uncertainty regarding the true state of the system ~\cite{cassandra1998survey,bravo2019use}. Solving POMDPs is computationally challenging, leading to the development of various solvers, including Monte-Carlo tree search~\cite{katt2017learning}, reinforcement-learning variants~\cite{singh2021structured}, and diverse approximation schemes~\cite{kearns1999approximate}.  

Many application domains share a \emph{monotonic} structure, where the latent state of individual components degrades stochastically over time unless a costly restorative action is taken.  
Canonical examples include online advertising ~\cite{boutilier2016budget}, inventory replenishment~\cite{shin2015mdp}, and sequential repair or maintenance scheduling for physical assets~\cite{miehling2020monotonicity,bhattacharya2021multiagent}.  
\begin{figure}[!htb]
    \centering
    \includegraphics[width=0.9\textwidth]{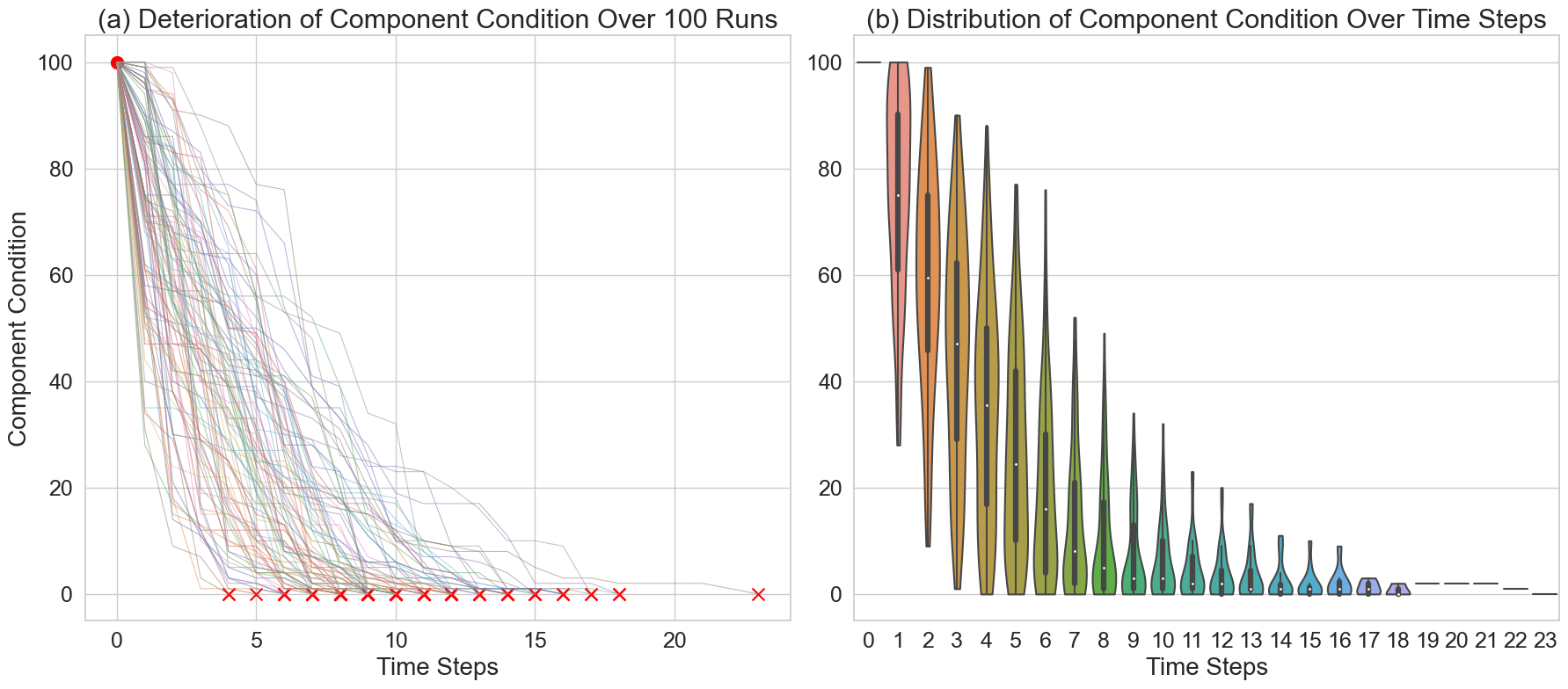}
    \caption{Condition of infrastructure component over time. (a) Line plot showing component condition over time for 100 runs. The red x marks denote the time step when condition reaches 0. (b) Violin-plot showing distribution of component condition for different time steps.}
    \label{fig:condition}
\end{figure}
Figure~\ref{fig:condition} shows this stochastic decline and the
probability distribution of a sample component’s condition at successive
time steps. While prior work, such as \cite{bhattacharya2020reinforcement}, has addressed optimal policies for single-component
systems, real-world systems---from building portfolios to exchange-traded-fund
(ETF) baskets---naturally involve \emph{many} such components
\cite{daulat2024challenges}.

In this paper, we address the challenge of computing approximately optimal policies for budget-constrained multi-component monotonic POMDPs. We assume that each component POMDP operates independently in terms of transition probabilities, but they are collectively constrained by the shared budget. Substantial work has been done to solve budget-constrained POMDPs \cite{lee2018monte,undurti2010online,khonji2019approximability}. However, the complexity of these algorithms is often exponential in the number of states of a single POMDP. For a multi-component POMDP, where the overall state space is the Cartesian product of individual component state spaces, this complexity consequently becomes exponential in the number of components. Thus, these methods become computationally intractable for multi-component POMDPs with a large number of components. A key challenge in solving budget-constrained multi-component POMDPs is how to optimally allocate the shared budget across the multiple components. In \citet{vora2023welfare}, the authors propose a welfare-maximization method for solving budget-constrained multi-component POMDPs. However, the method in that paper requires generating optimal policies for multiple budget values for every component POMDP to get the optimal budget allocation. Hence, it cannot be scaled to a large number of components.
% \paragraph{Our insight.}  
% We first show that the optimal value function of a \emph{single-component} monotonic POMDP is \textbf{concave in the available budget}.  
% This \emph{budget-concavity} result (proved in full in Section \ref{subsec:concavity}) enables an efficient decomposition strategy:
% \begin{enumerate*}[label=(\arabic*)]
% \item allocate budget across components by maximising a concave surrogate built from random-forest regressors, and
% \item learn a near-optimal policy for each component–budget pair with an \emph{oracle-guided, meta-trained Proximal Policy optimization (PPO)} agent whose oracle comes from value iteration on the fully observable counterpart.
% \end{enumerate*}

\paragraph{Our insight.}  
The primary computational bottleneck in solving budget-constrained multi-component POMDPs is the \emph{coupling} induced by the shared budget.  
If that budget could be split \emph{a-priori}, the joint POMDP would
factor into \(n\) independent, single-component problems solvable in
parallel. To enable this decomposition, we prove that the optimal value function of a \emph{single}
monotonic POMDP is \textbf{concave} in its allocated budget.  
This budget-concavity lets us decouple first, optimise second:

\begin{enumerate}[label=(\arabic*)]
\item \emph{Budget allocation.}  We maximize a concave surrogate of the value function, estimated with a random-forest regressor,
      to distribute the global budget across components; and
\item \emph{Component policies.}  With budgets fixed, we learn a
      near-optimal policy for each component–budget pair using an
      \emph{oracle-guided, meta-trained} Proximal Policy Optimization
      (PPO) agent, where the oracle is obtained by value iteration on the
      fully observable counterpart.
\end{enumerate}

The result is a scalable solution whose runtime grows linearly with the
number of components while retaining strong performance guarantees.

% \paragraph{Broader scope.}  
% Although motivated by infrastructure maintenance, this two-step scheme applies to other monotonic, budget-limited settings.  
% We demonstrate this generality on a \emph{budget-constrained ETF portfolio-selection} task inspired by~\cite{cho2017robust}: each asset’s latent “market sign’’ (down/flat/up) drifts unfavourably unless capital is re-allocated.  
% Both domains showcase thousands of components (infrastructure elements or daily trading decisions), emphasising scalability.

\paragraph{Contributions.}
\begin{enumerate}
    \item \textbf{Theory.}  We prove budget-concavity of the optimal value function for monotonic POMDPs. While prior works implicitly assume and use this budget-concavity, our work provides the first general structural guarantee that formally links budget availability to expected return.
    \item \textbf{Algorithms.}  We introduce (i) a random-forest budget-allocation module that exploits concavity for fast global optimization, and (ii) an oracle-guided meta-PPO solver for each single-component POMDP.
    \item \textbf{Empirical evidence.}  On two domains---preventive maintenance of a 1000-component building and portfolio loss-budget management with recapitalization---we outperform baseline heuristics and vanilla PPO, whilst solution time of the proposed approach scales \emph{linearly} in the number of components.
    \item \textbf{Complexity analysis.}  We provide a detailed runtime study confirming linear growth in wall-clock time as components increase from \(n=10\) to \(n=1000\).
\end{enumerate}

The remainder of the paper is organized as follows.  
Section~\ref{sec:related} surveys related work on budget-constrained POMDPs and large-scale maintenance or portfolio problems.  
Section~\ref{sec:problem} formalises the budget-constrained multi-component monotonic POMDP.  
Section~\ref{sec:method} presents our solution pipeline:  
\textbf{(i)} Subsection~\ref{subsec:concavity} proves budget–concavity of the single-component value function;  
\textbf{(ii)} Subsection~\ref{subsec:allocation} exploits this structure to allocate the global budget via a random-forest surrogate; and  
\textbf{(iii)} Subsection~\ref{subsec:rl} derives an oracle-guided meta-PPO policy for each component and composes them into the overall controller.  
Section~\ref{sec:experiments} reports empirical results on infrastructure maintenance and financial loss-budget management, and  
Section~\ref{sec:conclusion} concludes with key findings and future directions.

\section{Preliminaries and Related Work}\label{sec:related}
\subsection{Partially Observable Markov Decision Processes}
A discrete-time finite-horizon Partially Observable Markov Decision Process (POMDP) \cite{pomdpcassandra} $M$ is defined by the 7-tuple $(\mathcal{S}, A, T, \Omega, O, R, H)$, which denotes the state space, action space, state transition function, observation space, observation function, reward function and planning horizon, respectively.
% In the context of a POMDP, at each time step $k$, the environment resides in some state $s_k \in \mathcal{S}$, and the agent interacts with the environment by taking an action $a_k \in A$. This action results in the environment transitioning to a new state $s_{k+1} \in \mathcal{S}$ in the subsequent time step with probability $T(s_k, a_k, s_{k+1})$. Simultaneously, the agent receives an observation $o_k \in \Omega$ regarding the next state of the environment with probability $O(o_k, s_{k+1}, a_k)$, which depends on the new state and the action taken by the agent.
In a POMDP, the agent does not have direct access to the true state of the environment. Instead, the agent may maintain a \textit{belief state}, representing a probability distribution over $\mathcal{S}$. This belief is updated based on the received observation using Bayes' rule \cite{araya2010pomdp}.

% The objective of optimal policy synthesis for a finite-horizon POMDP is to determine a sequence of actions that maximizes the expected total reward over the planning horizon $H$.

\subsection{POMDP Solution Methods}
Computing optimal policies for a POMDP is generally PSPACE-complete \cite{mundhenk2000complexity,vlassis2012computational}. Thus, to address the computational intractability of solving POMDPs, various approximation methods have been widely used \cite{poupart2002value,pineau2003point,roy2005finding}. Several reinforcement learning approaches have also been developed for computing approximate POMDP solutions \cite{azizzadenesheli2016reinforcement,igl2018deep}. However, these methods become computationally intractable when faced with the high dimensionality and shared resource constraints of budget-constrained multi-component monotonic POMDPs such as those considered in this paper. 

\subsection{Consumption MDPs and Budgeted POMDPs}
The integration of budget or resource constraints into Markov Decision Processes (MDPs) has been previously 
studied under the frameworks of Consumption MDPs \cite{blahoudek2020qualitative} and Budgeted POMDPs \cite{vora2023welfare}. However, the algorithm proposed in \citet{blahoudek2020qualitative} assumes full observability of the state and hence cannot be applied to budget-constrained POMDPs. A solution for budget-constrained multi-component POMDPs is presented in \citet{vora2023welfare}. However, the method in this paper requires repeated computations of optimal policies for different budget values for all component POMDPs and hence is not scalable to a budget-constrained multi-component POMDP with a large number of components. 

\begin{figure*}[th]
    \centering
    \includegraphics[width=\textwidth]{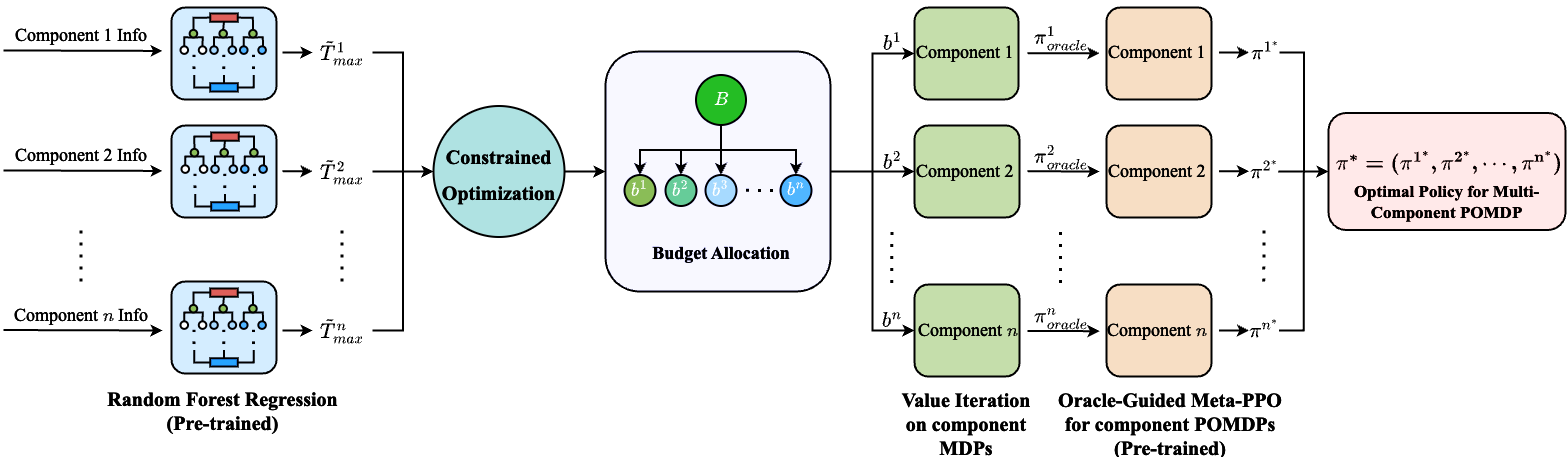}
    \caption{Architectural overview of the proposed approach.}
    \label{fig:arch}
\end{figure*}
\section{Problem Formulation}\label{sec:problem}
In this paper, we consider a weakly-coupled multi-component monotonic POMDP with a total budget. A weakly-coupled multi-component POMDP refers to a system where the individual component POMDPs have independent transition probabilities but are interconnected through a shared budget $B$. This shared budget introduces a weak coupling between the components, as the allocation of budget to one component affects the available budget for the others. The state space for an $n$-component monotonic budget-constrained POMDP is given by
$\mathcal{S} = \prod_{i=1}^n \mathcal{S}_i,$
where $\mathcal{S}_i$ represents the state space for component $i$, and $i \in \{1, \ldots, n\}$. The action space is given by
$\mathcal{A} = \prod_{i=1}^n \mathcal{A}_i,$
where the action space for component $i$ is $\mathcal{A}_i = \{d^i, q^i, m^i\}$. Each action incurs a fixed cost. The state at time instant $k$ is an $n$-tuple, $s_k = (s_k^1, s_k^2, \cdots, s_k^n),$ where $s_k^i \in \mathcal{S}_i = \{0, 1, \ldots, \bar{s}\}$ denotes the state of component $i$, and $\bar{s} \in \mathbb{N}_0$ is the maximum possible value of $s_k^i$. Here, $\mathbb{N}_0$ denotes the set of non-negative integers. Similarly, the action at time $k$ is given by $a_k = (a_k^1, a_k^2, \cdots, a_k^n)$ and the cost associated with this action is given by $c_{a_k} = \sum_{i=1}^n c_{a^i_k},$ where $c_{a^i_k}$ represents the cost associated with each action $a_k^i$.
The transition function for the multi-component POMDP is:
$$T(s_k, a_k, s_{k+1}) = \prod_{i=1}^n T_i(s^i_k, a^i_k, s^i_{k+1}).$$ 
The transition probability function for each component $i$ is:
\begin{equation}
    T^i(s^i_k,a^i_k,s^i_{k+1}) = 
    \begin{cases}
    p_1^i(s^i_k,a^i_k,s^i_{k+1}), & \text{if } a^i_k = m^i\text{ and } \\
    &  s^i_k \leq s^i_{k+1} \leq \bar{s},\\
    p_2^i(s^i_k,a^i_k,s^i_{k+1}), & \text{if }a^i_k \in \{d^i, q^i\}\\
    & \text{and }s^i_{k+1} \leq s^i_k,\\
    1, & \text{if }a^i_k \in A^i\text{ and }\\
    & s^i_{k+1} = 0 = s^i_k ,\\
    0, & \text{otherwise}.
    \end{cases}\label{eq:transition}
\end{equation}
Here, action $m^i$ is a restorative action that increases the state value, with the increase being upper bounded by $\bar{s}$. In contrast, actions $d^i$ and $q^i$ decrease the state value. Moreover, $s_k^i = 0$ is an absorbing state for all $k,i$. Finally, the observation probability function for each component follows the model from \citet{vora2023welfare}, where action $q_i$ gives true state information and the other two actions provide no information about the true state.
\subsection{Problem Statement}
The primary objective of this paper is to determine a policy $\pi^*$ for this multi-component monotonic POMDP over a horizon $H$, that maximizes the sum of expectations of the individual times before reaching the absorbing state for each component, while adhering to the total budget $B$. We denote this maximal time $k$ by $T_{max} = \sum_{i=1}^n T^i_{max}$, where $T^i_{max}$ denotes the corresponding maximal time for component $i$. Mathematically, the problem can be formulated as:
\begin{equation}
\begin{aligned}
    \max_{\pi}\left( \sum_{i=1}^n \mathbb{E}[T^i_{max}(\pi)]\right) \\
    \text{s.t.} \sum_{k=0}^H c_{a_k}(\pi) \leq B.\label{eq:formulation}
\end{aligned}
\end{equation}
In this formulation, $\pi$ represents the policy, and both $T^i_{max}$ and $c_{a_k}$ depend on $\pi$. For simplicity, we will not explicitly denote this dependence in the remainder of this paper.
There are many other possible formulations of the objective of the problem statement like a \textit{maxmin} formulation:
\begin{equation}
\max_{\pi} \min_{i} \mathbb{E}[T^i_{max}(\pi)].\label{eq:altformulation}
\end{equation}
In this paper we consider the formulation given by \eqref{eq:formulation}.

% \section{Solution Approach}\label{sec:method}
% In this section, we present our methodology for solving a budget-constrained multi-component monotonic POMDP. Figure~\ref{fig:arch} presents an architectural overview of our proposed approach. We first input information about all $n$ component POMDPs into a pre-trained random forest regressor to get the $T^i_{max}$ for each component $i$ as a function of the budget allocated to the component. Next, we propose an appropriate allocation of the shared budget $B$ among the component POMDPs by solving a constrained maximization problem. We then compute the oracle policy for each component POMDP and budget pair through value iteration applied to the corresponding component MDP. Finally, using these oracle policies and a meta-trained reinforcement learning (RL) agent, we solve each component POMDP with respect to the allocated budget and consequently propose a policy for the multi-component POMDP. Note that an alternate allocation strategy could involve redistributing the budget at every time step during planning. However, such a method would be computationally more expensive, due to the repeated computation of the allocation, as compared to our proposed a priori budget distribution approach.

\section{Solution Approach}\label{sec:method}

In this section, we present our methodology for solving a budget-constrained multi-component monotonic POMDP. Figure~\ref{fig:arch} presents an architectural overview of our proposed approach.  
The key idea is to \emph{decouple first, optimize second}.  
Allocating the shared budget \emph{as a first step of} planning shrinks the original large joint POMDP into \(n\) independent single-component POMDPs. Each of these single-component POMDPs now operates with its own fixed budget cap, which is determined by the initial allocation. This transformation converts a problem that is intractable for \(n\!\gg\!1\) into \(n\) modest ones that can be solved in parallel. We organize the section accordingly:
\begin{itemize}[leftmargin=*]
  \item \textbf{Structural Result: Budget Concavity}
        (Section~\ref{subsec:concavity}): We prove that each component’s value
        function is concave in its budget.
  \item \textbf{Stage~1: Budget Allocation}
        (Section~\ref{subsec:allocation}): Leveraging concavity, we fit a
        random-forest surrogate of the value function and solve a tractable
        concave maximization problem to distribute the shared global budget across components.
  \item \textbf{Stage~2: Oracle-Guided Meta-PPO}
        (Section~\ref{subsec:rl}): With budgets fixed, we learn near-optimal
        policies for each component–budget pair (with respect to that component's allocated budget and local POMDP) using an
        oracle-guided, meta-trained PPO agent, then compose these into the
        overall multi-component policy.
\end{itemize}
Note that an alternate allocation strategy could involve redistributing the budget at every time step during planning. However, such a method would be computationally more expensive than our proposed approach due to the repeated computation of the allocation.

\subsection{Budget–Concavity of the Value Function}\label{subsec:concavity}

We first show that the optimal value function of a single-component belief-MDP---derived from a monotonic POMDP with random, nonnegative action costs---is concave in the budget variable $B$ for any finite planning horizon $H\geq0$.

% \vspace{0.8em}
\subsection*{Setting}

\noindent
We consider a POMDP defined by the tuple $\langle S, A, T, R, \Omega, O, \gamma, \mathcal{C} \rangle$, where:
\begin{itemize}
    \item $S$ is a finite state space and $A$ is a finite action space, as defined in Section~\ref{sec:problem}.
    \item $T(s'|s,a)$ is the transition kernel \eqref{eq:transition}; $R(s,a)$ is the reward function.
    \item $\Omega$ is the observation space; $O(o | s', a)$ is the observation model as defined in Section~\ref{sec:problem}.
    \item $\gamma \in [0, 1)$ is the discount factor.
    \item $\mathcal{C}(s,a)$ is the distribution of a random, nonnegative cost incurred by taking action $a$ in state $s$. The specific cost, $c$, is drawn from this distribution.
\end{itemize}

\noindent
Following standard practice in POMDP literature \cite{cassandra1998survey}, we reformulate this POMDP as a belief-MDP with state $(b, B)$, where $b \in \Delta(S)$ is the belief (posterior distribution over hidden states), and $B \geq 0$ is a remaining budget. The reward at belief $b$ under action $a$ is:
\[
\rho(b, a) := \sum_{s \in S} b(s) R(s, a),
\]
and the budget evolves as $B \mapsto B - c$, where $c$ is a realization from the random cost $C(s, a)$ under the current belief. To prove the budget-concavity of the value function for this belief-MDP, we first establish two foundational properties concerning concavity. These lemmas demonstrate how concavity is preserved under common mathematical operations relevant to dynamic programming.

\begin{lemma}[Concavity under Affine Shift] If $f(x)$ is concave on an interval $I$, then $f(x - l)$ is concave on the interval $\{y \mid y = x+l, x \in I\}$ for any constant $l$.
\end{lemma}
\begin{proof}
    This lemma is a standard result in convex analysis \cite{boyd2004convex}.
\end{proof}

\begin{lemma}[Expectation Preserves Concavity]
Let $f(B, \xi)$ be concave in $B$ for every realization $\xi$. If $\xi$ is a random variable following an arbitrary probability distribution, then $\mathbb{E}_\xi[f(B,\xi)]$ is also concave in $B$.
\end{lemma}

\begin{proof}
    Fix any $B_1, B_2 \in \mathbb{R}$ and any $\lambda \in [0,1]$. Let $g(B) = \mathbb{E}_\xi[f(B, \xi)]$. Then
\[
g(\lambda B_1 + (1 - \lambda) B_2) = \mathbb{E}_\xi\left[ f\big(\lambda B_1 + (1 - \lambda) B_2, \xi\big) \right].
\]
Since $f(B, \xi)$ is concave in $B$ for each $\xi$, we have
\[
f\big(\lambda B_1 + (1 - \lambda) B_2, \xi\big) \geq \lambda f(B_1, \xi) + (1 - \lambda) f(B_2, \xi)
\]
for all $\xi$. Taking expectations on both sides yields
\[
\mathbb{E}_\xi\left[ f(\lambda B_1 + (1 - \lambda) B_2, \xi) \right] \geq \lambda \mathbb{E}_\xi\left[ f(B_1, \xi) \right] + (1 - \lambda) \mathbb{E}_\xi\left[ f(B_2, \xi) \right],
\]
that is,
\[
g(\lambda B_1 + (1 - \lambda) B_2) \geq \lambda g(B_1) + (1 - \lambda) g(B_2).
\]
Thus, $g$ is concave in $B$.
\end{proof}
Having established these fundamental properties regarding the preservation of concavity under affine shifts and expectations, we will now prove that the optimal value function of a monotonic POMDP is concave with respect to the available budget.
\begin{theorem}[Budget Concavity]\label{thm:concave}
For any fixed belief $b \in \Delta(S)$ and horizon $H \geq 0$, the value function $V_H(b, B)$ is concave in $B$ on $[0, \infty)$.
\end{theorem}

\vspace{1em}
\noindent
\begin{proof}
We proceed by mathematical induction on $H$.

\vspace{0.5em}
\noindent
{\bf Base Case ($H = 0$).} At horizon zero, there are no rewards:
\[
V_0(b, B) = 0 \quad \text{for all } b \in \Delta(S),\ B \geq 0.
\]
\noindent
Function $V_0$ is thus trivially concave.

\vspace{0.8em}
\noindent
{\bf Inductive Hypothesis.} Suppose that for some $H \geq 0$, the function $V_H(b, B)$ is concave in $B$ for every belief $b$.

\vspace{0.8em}
\noindent
{\bf Inductive Step.} We aim to prove that $V_{H+1}(b, B)$ is concave in $B$ for all $b$. The Bellman equation in the belief-MDP is:
\[
V_{H+1}(b, B) = \max_{a \in A} \left\{ \rho(b, a) + \gamma\, \mathbb{E}_{o, c \mid b,a} \left[ V_H(b', B - c) \right] \right\},
\]
where $b'$ is the updated belief after taking action $a$ and observing $o$. The expectation $\mathbb{E}_{o, c \mid b,a}$ is taken over the random observation $o$ and cost $c$ given the current belief $b$ and chosen action $a$.

\noindent
Define the inner expectation as:
\[
g(a, b, B) := \mathbb{E}_{o,c \mid b,a} \left[ V_H(b', B - c) \right].
\]
Apply Lemma 1 and inductive hypothesis to assert that $B \mapsto V_H(b', B - c)$ is concave for each $(o, c)$. Then Lemma 2 implies that $g(B)$, being the expectation over such functions, is also concave.

\noindent
Therefore, the $Q$-value
\[
Q_{H+1}(b, B, a) := \rho(b, a) + \gamma g(B)
\]
is concave in $B$ for each $a$.

\noindent
Finally, the value function is
\[
V_{H+1}(b, B) = \max_{a \in A} Q_{H+1}(b, B, a),
\]
which is the pointwise maximum of finitely many concave functions, and hence concave itself.
\end{proof} 

\noindent
\subsubsection{Relating \texorpdfstring{$\mathbb{E}[T_{\max}]$}{E[Tmax]} to the Value Function}
We proved the budget-concavity of the value function $V_H(s,B)$ in Theorem~\ref{thm:concave}. In our problem setting \eqref{eq:formulation}, however, we aim to maximize the expected time to failure $E[T_{\max}]$. Specifically, in many practical applications such as preventive maintenance or portfolio management, the objective can be naturally framed as maximizing the expected time until a critical failure occurs or a budget is exhausted. We now show how the concavity property extends to $E[T_{\max}]$, which serves as the objective function for our initial budget allocation stage.
\begin{lemma}[Expected-time equivalence]\label{lemma:etmax}
Consider the reward function
\begin{equation}
R(s,a)=
\begin{cases}
1,& s\neq0,\\
0,& s=0,
\end{cases}
\label{eq:reward_proof}
\end{equation}
. Let \(V(s,B)\) be the corresponding optimal value function.  Denote by
\(\mathbb{E}[T_{\max}(B)]\) the expected time to reach the absorbing
state \(s=0\) under the optimal budget-feasible policy.  Then
\[
V(s,B)=\mathbb{E}[T_{\max}(B)].
\]
\end{lemma}

\begin{proof}
Under reward scheme \eqref{eq:reward_proof} each non-absorbing step contributes
exactly \(1\) to the return; steps in state \(0\) contribute \(0\).
Hence, for any budget-feasible policy \(\pi\),
\[
\text{Total reward}=\mathbb{E}\Bigl[\sum_{t=0}^{H}\mathbf{1}\{s_t\neq0\}\Bigr]
=\mathbb{E}[T_{\max}(\pi)].
\]
Taking the maximum over all budget-feasible policies yields
\(V(s,B)=\mathbb{E}[T_{\max}(B)]\).
\end{proof}

\begin{corollary}\label{cor:etmax_concave}
Because \(B \mapsto V(s,B)\) is concave by Theorem~\ref{thm:concave},
the expected absorption time \(\mathbb{E}[T_{\max}(B)]\) is also concave
in the allocated budget \(B\).
\end{corollary}

\subsection{Random Forest Approach for Optimal Budget Allocation}\label{subsec:allocation}
By Corollary~\ref{cor:etmax_concave}, the expected maximal survival time
$\mathbb{E}[T_{\max}(B)]$ is a concave function of the budget allocated
to a single component.  This structural property lets us treat budget
splitting across $n$ components as a \emph{concave maximization}
problem---one that is both tractable and amenable to surrogate modeling.
Each component evolves independently but competes for the shared budget, rendering the components weakly coupled. While reinforcement learning algorithms have made significant advances, they often face challenges when scaling to the extremely large state and action spaces characteristic of multi-component systems \cite{sutton2018reinforcement}. To address this scalability issue, our remedy is an \emph{a-priori} budget distribution that decouples
the system.  Concretely, for component $i$ we approximate the concave map
$B\mapsto\mathbb{E}[T_{\max}^i(B)]$ by the exponential surrogate  
\begin{equation}
    \widetilde{T}^i_{\max}(B)
    \;=\;
    \alpha^{i}\,e^{\beta^{i}B} + \gamma^{i},
    \label{eq:ttf_budget}
\end{equation}
where $(\alpha^{i},\beta^{i},\gamma^{i})$ are constants. While many other concave functions could be used to model $\Tilde{T}^i_{\max}$, we empirically observe that the exponential function provides a good fit for the data (see Appendix~\ref{appdx:fun_approx}). We use a random forest regressor \cite{breiman2001random} to estimate the parameters of this exponential function. The training dataset for this model is obtained via non-linear least squares regression on multiple $(\mathbb{E}[T_{\max}], b)$ pairs for various budget-constrained single-component monotonic POMDPs. The input to this model includes specific statistics related to the POMDP's transition function, which are the expected time to reach state 0 without repairs, $\mathbb{E}[T]$, and the variance of this expected time, $\sigma^2_{\mathbb{E}[T]}$, as well as the various actions costs.

Let $b^{i}$ denote the budget assigned to component $i$ and
$\widetilde{T}^i_{\max}$ its surrogate survival time.  The allocation
problem becomes
\begin{equation}
    \begin{aligned}
        \max_{b^{1:n}}\;&\sum_{i=1}^{n}\widetilde{T}^i_{\max}(b^{i})
        \\
        \text{s.t. }\;&\sum_{i=1}^{n} b^{i}\;\le\;B,\qquad
                      b^{i}\;\ge\;0\;\;\forall i,
    \end{aligned}
    \label{eq:maximize}
\end{equation}
a concave maximization with linear constraints.  
Because each surrogate in~\eqref{eq:ttf_budget} is concave, the problem is
globally tractable and we solve it with off-the-shelf convex
optimizers. Solving \eqref{eq:maximize} yields the approximately optimal budget allocation among the individual components. The next subsection shows how an oracle-guided meta-PPO agent learns the
individual component policies given this budget allocation.

\subsection{Oracle-Guided RL for a Budget-Constrained Single Component}\label{subsec:rl}

Given the per-component budgets $b^i$ obtained in
Section~\ref{subsec:allocation}, we now derive a near-optimal control
policy for each single-component budget-constrained monotonic POMDP.  We
adopt the budget-augmented POMDP (bPOMDP) formalism of
\citet{vora2023welfare}, in which the state includes an additional,
fully-observable coordinate that tracks cumulative cost.

The oracle policy is denoted as $\pi_{\text{oracle}}$ and is obtained by solving the corresponding MDP using value iteration. For a single-component monotonic POMDP with budget $B$, the corresponding MDP has an action space $\mathcal{A}_{\text{MDP}} = \{d, m\}$, identical transition probabilities as the POMDP, and \textit{full observability of the state}.  We then train a Proximal Policy
optimization (PPO) agent~\cite{schulman2017proximal} that \emph{queries}
this oracle selectively: at each time step it chooses either to inspect
($q$) or to defer ($\neg q$), in which case the action recommended by the
oracle is executed. Since the full state is not observable in a POMDP, we utilize the belief $b_s$ for planning. The agent's belief of the true state is updated at each time step using a particle filter approach. For our work, we empirically observe that using the expected belief $\bar{b_s}$ and the variance of the belief $\sigma^2_{b_s}$ suffices for planning. 

Hence, for the proposed oracle policy-guided PPO agent, the state at time instant $k$ is given by the vector $[\bar{b}_{s_k}, c_k, \sigma^2_{b_{s_k}}]$. Furthermore, the reward function is defined as follows:
\begin{equation*}
    R(s_k, c_k, a_k) = 
    \begin{cases}
        r_1 < 0, & \text{if } c_k > B, \\
        r_2 < 0, & \text{if } \left\lfloor{\bar{b}_{s_k}}\right\rfloor = 0, \\
        r_3 = \frac{k}{H} - \alpha |\bar{b}_{s_k} - s_k|, & \text{if } \bar{b}_{s_k}, c_k > 0,
    \end{cases}
\end{equation*}
where $|r_1| > |r_2| > |r_3|$ for all $k$, $0 < \alpha < 1$ and $\left\lfloor{.}\right\rfloor$ denotes the floor function. This reward function imposes substantial negative rewards for exceeding the budget $B$ and allowing the state $s_k$ to reach 0. Additionally, at each time step, the agent receives a positive reward proportional to the time step for maintaining $s_k$ above zero and incurs a penalty proportional to the absolute error between the expected belief and the true state. As a result, the agent gets higher rewards for keeping $s_k>0$ for a longer time and is heavily penalized when the expected belief deviates significantly from the true state. It is crucial to note that during training, the agent relies solely on the observed reward signals, without access to the true state.

\subsection{Optimal Policy for Multi-Component Monotonic POMDPs}
We now integrate the approaches described in Section~\ref{subsec:allocation} and Section~\ref{subsec:rl} to compute the optimal policy for an $n$-component POMDP, where $n$ is substantially large. Utilizing the random forest regressor, we efficiently approximate $\mathbb{E}[T_{\max}]$ for each component $i$. Additionally, we meta-train the oracle-guided PPO agent by continuously updating the policy network's parameters over a randomly selected subset of components and budget values. This approach allows the agent to generalize across components.
This meta-trained agent is then utilized to derive the optimal policy $\pi^{i^*}$ for each component $i$, following the optimal budget allocation obtained from \eqref{eq:maximize}. Consequently, the overall policy for the multi-component POMDP is:
\begin{equation*}
    \pi^*(s_k, a_k) = (\pi^{1^*}(s_k^1, a_k^1), \pi^{2^*}(s_k^2, a_k^2),\cdots, \pi^{n^*}(s_k^n, a_k^n)).
\end{equation*}
While this policy is not guaranteed to be globally optimal for the entire multi-component POMDP, we empirically observe that it performs well in practice while respecting the budget constraints. We validate this approach by evaluating its performance on real-world data in the subsequent section.

\section{Implementation and Evaluation}\label{sec:experiments}
In this section, we empirically validate our proposed framework on two disparate domains. The first domain, which we call the \emph{infrastructure scenario}, involves preventive maintenance for a large‐scale building comprising 1000 independent components whose latent condition stochastically degrades over time; our goal is to allocate a finite maintenance budget to maximize the expected survival time of all components. The second domain, the \emph{financial loss‐budget scenario}, addresses portfolio risk management using daily price data for S\&P 500 constituents, where each asset is endowed with a debit‐only loss budget that depletes under negative returns and can be replenished only through costly recapitalization. In the infrastructure scenario, we compare our oracle‐guided meta‐PPO approach against baseline heuristics, vanilla PPO, and an idealized oracle policy, reporting results on survival time, cost efficiency, and computational scalability across a range of budget levels. In the financial loss‐budget scenario, we focus on analyzing the learned recapitalization policy and assessing the generalizability and window robustness of proposed oracle‐guided meta‐PPO.

\subsection{Implementation and 
Evaluation for Infrastructure Scenario}
In this section, we evaluate the efficacy of the proposed methodology for determining the optimal policy for a very large multi-component budget-constrained POMDP. Specifically, we compare our approach against existing methods in the context of a multi-component building maintenance scenario managed by a team of agents. We also perform a computational complexity analysis of the proposed approach, for varying number of components.

We consider an administrative building comprising 1000 infrastructure components, including roofing elements, water fountains, lighting systems, and boilers. Each component's health is quantified by the Condition Index (CI) \cite{grussing2006condition}, which ranges from 0 to 100. For each infrastructure component, we utilize historical CI data to generate the transition probabilities for the corresponding POMDP, modeled using the Weibull distribution \cite{grussing2006condition}. We use the \texttt{weibull\_min} class from the \texttt{scipy.stats} module in Python to simulate the CI transitions over time. While a seed can be set using the \texttt{random\_state} parameter in \texttt{weibull\_min} for reproducibility, we did not set one to preserve the stochastic nature of the CI transitions. The condition index deteriorates stochastically over time, influenced by various factors, and can only be accurately assessed through explicit inspections, which incur a cost. A component is considered to have failed when its CI falls below a failure threshold, which is assumed to be 0. Components can be repaired to increase their CI. The building is allocated a maintenance budget of $B =$ 500,000 units for a given horizon of 100 decision steps. At the beginning of the horizon, the CI of all components is 100. The objective of the agents is to maximize the time until failure of the components by efficiently allocating the budget among the components and performing repairs and inspections as needed. The replacement costs (ranging from 50 to 500 units) and inspection costs (ranging from 1 to 5 units) of these components are derived from industry averages. Consistent with the approach described in Section~\ref{subsec:rl}, we model this objective as a POMDP (with $\alpha=10^{-3}$ in the reward function). This POMDP has roughly $10^{2000}$ states and $3^{1000}$ actions.

\begin{figure*}[th]
    \centering
    \includegraphics[width=\textwidth]{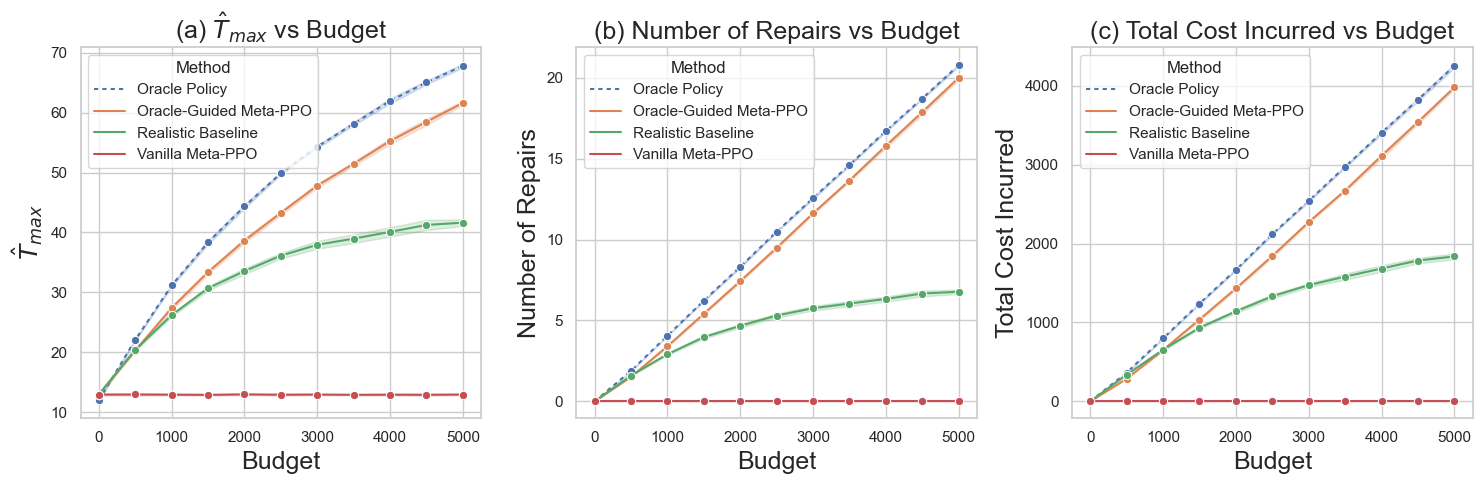}
    \caption{Performance comparison of oracle policy, oracle-guided meta-PPO, realistic baseline and vanilla meta-PPO. (a) Comparison of $\hat{T}_{max}$ values for all 1000 components across different budget values allocated to each component. (b) Comparison of average number of repairs performed by the agent under each of the four policies. (c) Comparison of average total cost incurred by the agent over the planning horizon for each of the four policies.}
    \label{fig:compare_metrics}
\end{figure*}
\subsubsection{Analysis of Maintenance Policy} \label{subsec:maintenance}
We begin by analyzing the performance of the maintenance policy derived using the proposed oracle-guided meta-PPO strategy for a single-component POMDP representing a component $i$ of the 1000 components. This policy is compared with the performance of the oracle policy on the corresponding component MDP. Since the oracle policy has full observability of the state, it is expected to always perform better than the proposed approach. Additionally, we evaluate two baseline policies:
\begin{enumerate}
    \item A heuristic policy often used in practice \cite{lam1994optimal,straub2004generic} where the agent performs inspections at regular intervals and repairs the component when its expected belief about the Condition Index (CI) falls below a predefined threshold. We chose an inspection interval of 5 steps and a repair threshold of 15 after extensive experiments with intervals ranging from 1 to 10 steps and repair thresholds from 5 to 50.
    \item A vanilla meta-PPO agent that is trained on the same subset of component-budget pairs as the oracle-guided agent, but without an oracle policy.
\end{enumerate}
Both the oracle-guided meta-PPO and vanilla meta-PPO are trained for 2M time steps each, with an Adam stepsize of $10^{-4}$, a minibatch size 128, policy update horizon of $T=4096$ and discount factor $0.95$. All other hyperparameters follow those used in \citet{schulman2017proximal}. We perform 100 simulations for this component to obtain the corresponding $T^i_{\max}$ values, which are then averaged over the runs for a given budget value allocated to the component. This process is repeated for all 1000 components and the run-averaged $T^i_{\max}$ values are then averaged across components. We compare this average denoted by $\hat{T}_{max}$ for 11 different budget values ranging from 0 to 5000 units, along with the average number of repairs performed by the agent and the average cost incurred over the planning horizon. Figure~\ref{fig:compare_metrics} illustrates a comparison of these metrics for all four policies. We observe that the proposed approach significantly outperforms the baselines. The oracle-guided meta-PPO agent nearly matches the performance of the oracle policy for all 3 metrics, presumably due to the low inspection costs of the components. If inspection costs were significantly higher, the agent's performance would likely diverge from the oracle policy, which is an expected outcome given the budget constraints. We also infer that the vanilla meta-PPO agent has only learnt to not violate the budget constraint by not performing any repairs. These results demonstrate the value of incorporating an oracle policy into the training of a reinforcement learning agent.
\begin{figure}[!htb]
    \centering
    \includegraphics[width=0.8\textwidth]{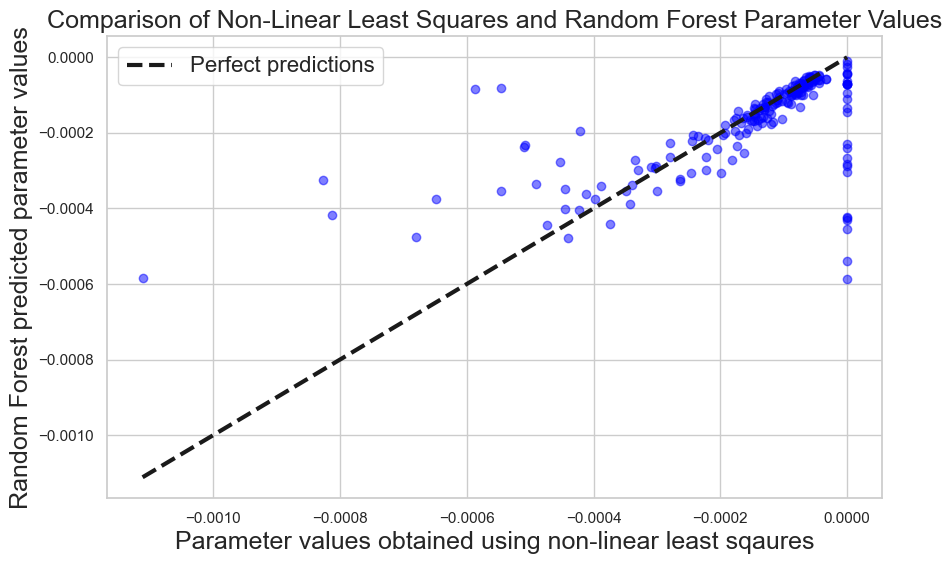}
    \caption{Performance of random forest model for predicting the value of parameter $\beta$ for a test dataset of 200 components. The horizontal axis represents parameter values obtained via non-linear least squares and vertical axis represents predicted values. The dotted line represents the $y=x$ line, i.e., perfect predictions.}
    \label{fig:compare_random_forest}
\end{figure}
\subsubsection{Analysis of Budget Allocation}
Next, we demonstrate the effectiveness of our random forest-based budget allocation strategy. We compare it with a baseline approach that allocates budgets proportional to the 
ratio of a component's replacement cost to its $\mathbb{E}[T]$. For a component $i$, we model its $\mathbb{E}[T^i_{\max}]$ using $\Tilde{T}^i_{\max}$ as given in \eqref{eq:ttf_budget} (see Appendix~\ref{appdx:fun_approx} for justification of this exponential form). 
The parameters $\alpha^i$ and $\gamma^i$ can be estimated directly by considering the boundary conditions: $\gamma^i$ is estimated by substituting $b^i = 0$, representing the scenario where no budget is available, and $\alpha^i$ is determined by substituting $b^i = \infty$, corresponding to the scenario of unlimited budget, where the supremum of $T^i_{\max}$ ($\sup_{b^i} T^i_{\max} = H = 100$) is reached.
We then train a random forest regressor to estimate parameter $\beta^i$. The training dataset is created by performing non-linear least squares regression on 11 distinct ($T^i_{max},b^i$) pairs each for 800 components. These pairs correspond to the run-averaged $T^i_{\max}$ values and the respective budget values $b^i$ from Section~\ref{subsec:maintenance}. The input to the random forest model is a vector consisting of the shape and scale factors of the Weibull distribution, which represent $\mathbb{E}[T]$ and $\sigma^2_{\mathbb{E}[T]}$, along with the replacement and inspection costs for a given component $i$. If a different distribution was used to model the transition probability, we would similarly extract the parameters, $\mathbb{E}[T]$ and $\sigma^2_{\mathbb{E}[T]}$, for inclusion in the input vector.
Figure~\ref{fig:compare_random_forest} shows the prediction performance of the random forest model for a test dataset of 200 components which were not encountered during training. We see that most points on the plot are very close to the perfect prediction line and bad predictions are few in number (29 out of 200 for error threshold of $10^{-4}$). The random forest model achieves a mean squared error (MSE) = $1.8\times10^{-8}$ for this test dataset. Note that the non-linear least squares regressor constrains $\beta^i$ to be $\leq 0$ and hence for some components we observe that $\beta^i=0$.
We use this trained random forest model to estimate $\Tilde{T}^i_{max}$ for all 1000 components. Finally, using these approximated expressions, we solve the constrained maximization problem described in \eqref{eq:maximize} to obtain the appropriate budget allocation for the components.
\begin{table}[!htb]
    \centering
    \begin{tabular}{c c}
        \toprule
        \textbf{Approach} & $T_{max}$ \\
        \midrule
        Random Forest Budget Allocation & 22,009.5 \\
        Baseline Budget Allocation & 16,445.4\\
        \bottomrule
    \end{tabular}
    \caption{$T_{max}$ (in steps), averaged over 100 runs, achieved under random forest and baseline budget allocations.}
    \label{tab:budget_allocation}
\end{table}
We quantify the performance of the random forest budget allocation and the baseline budget allocation algorithms by calculating the $T_{max} = \sum_{i} T^i_{max}$ and averaging it over 100 runs. For a fair comparison, these values are obtained using the oracle-guided meta-PPO approach for both allocation 
schemes. 

\begin{figure}[!htb]
    \centering
    \includegraphics[width=0.8\textwidth]{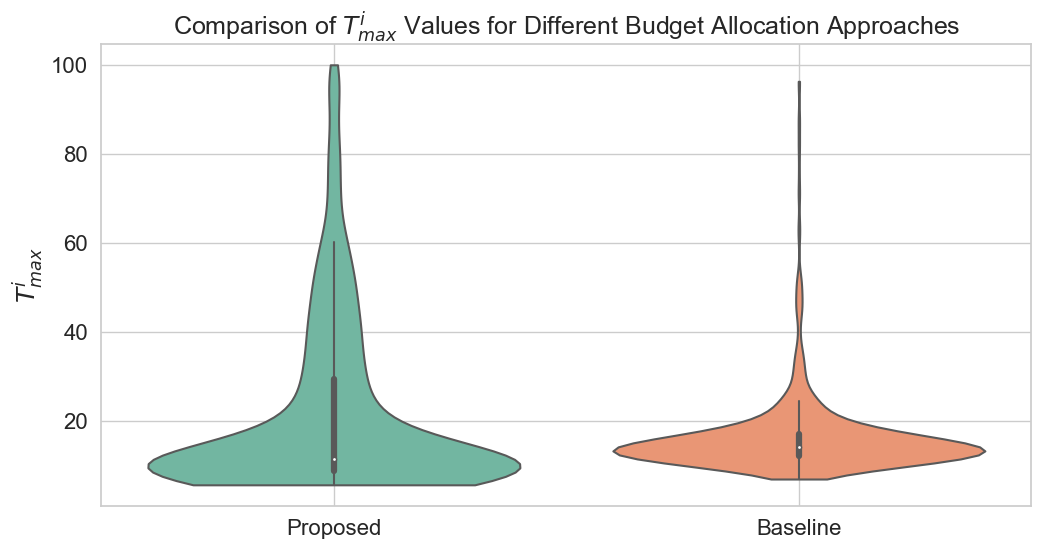}
    \caption{Performance comparison of random forest-based budget allocation and baseline budget allocation for all 1000 components for an overall budget of 500,000 units.}
    \label{fig:compare_budget_allocation}
\end{figure}

Table~\ref{tab:budget_allocation} shows the $T_{max}$ values achieved by both allocation approaches. 
The random forest budget allocation vastly outperforms the baseline approach.
Furthermore, Figure~\ref{fig:compare_budget_allocation} presents violin plots showing the distribution of the $T^i_{max}$ values achieved under the proposed and baseline budget allocations for all 1000 components. 
We observe that there are more components with higher $T^i_{max}$ values for the random forest budget allocation approach. 
Preliminary experiments on alternative objective formulations, such as the \textit{maxmin} approach given by \eqref{eq:altformulation}, also indicate that the proposed method consistently outperforms the baseline.
\subsubsection{Analysis of Time Complexity}
Finally, we analyze the time complexity of our proposed approach for varying number of components $N$. As mentioned earlier, our method comprises of four major steps:
\begin{enumerate}
    \item \textbf{Random Forest} regression for estimating $\Tilde{T}^i_{max}$ for each component $i$.
    \item \textbf{Budget Allocation} among components via constrained optimization.
    \item \textbf{MDP Value Iteration} for each component-budget pair to obtain the corresponding oracle policy.
    \item \textbf{Oracle-Guided Meta-PPO} to approximately solve each component POMDP.
\end{enumerate}
\begin{table}[!htb]
    \centering
    \begin{tabular}{c c c c c}
        \toprule
        \textbf{Number of Components} & \textbf{Random Forest} & \textbf{Budget Split} & \textbf{Value Iteration} & \textbf{Meta-PPO} \\
        \midrule
        1 & 0.9724 & 0.9046 & 113.7227 & 2.8885 \\
        2 & 0.8870 & 0.8314 & 116.3281 & 3.0858 \\
        5 & 0.8719 & 0.8207 & 135.3953 & 4.7940 \\
        10 & 0.8762 & 0.8132 & 280.4909 & 9.5495 \\
        20 & 0.9534 & 0.8997 & 451.2948 & 16.2575 \\
        50 & 0.9449 & 0.8916 & 1208.1000 & 33.7387 \\
        100 & 0.9324 & 0.9171 & 2389.5641 & 64.6809 \\
        500 & 0.9575 & 1.2226 & 10269.1037 & 313.9742 \\
        1000 & 0.9599 & 1.6232 & 20612.1734 & 627.7477 \\
        \bottomrule
    \end{tabular}
    \caption{Time taken (in seconds) for running each process with varying numbers of components, averaged over 10 runs.}
    \label{tab:time_taken}
\end{table}
\begin{figure}[!htb]
    \centering
    \includegraphics[width=0.8\textwidth]{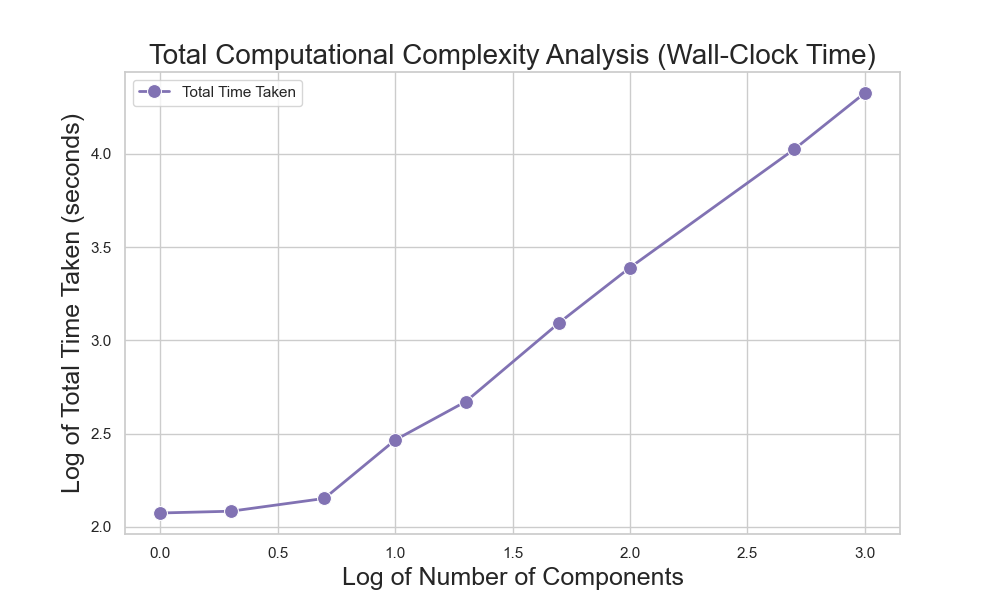}
    \caption{Log-log plot of computational complexity of the proposed approach for varying numbers of components.}
    \label{fig:time_complexity}
\end{figure}
Table~\ref{tab:time_taken} presents the times taken for running each of the four processes, with different number of components. The time complexity experiments were performed in Python on a laptop running MacOS with an M2 chip @3.49GHz CPU and 8GB RAM. The times taken for random forest and budget allocation steps are negligible compared to those for performing value iteration and generating optimal policies through meta-PPO. The value iteration is applied to each component independently and hence scales linearly with the number of components. Similarly, Step 4 involves applying the pre-trained policy to each component separately and thus is also linear in the number of components. Consequently, we expect that the time complexity of our algorithm is linear in the number of components, i.e., $O(n)$. This expectation is confirmed by the log-log plot of computational complexity shown in Figure~\ref{fig:time_complexity}. Our algorithm's performance is thus significantly faster as compared to existing POMDP solvers which would be exponential in the number of states and thus doubly exponential in the number of components \cite{silver2010monte}, \cite{pineau2003point}. If the problem is approached directly as a single POMDP, it will have a prohibitively vast state space of approximately $10^{2000}$ states. Previous work by \citet{vora2023welfare} demonstrated that standard methods indeed become computationally intractable after a few components due to this combinatorial explosion.

\subsection{Implementation and Evaluation for Financial Loss-Budget Management Scenario}

\noindent
Our second experimental scenario addresses a portfolio risk management task. We use daily price data for the S\&P 500 constituent stocks over a two-year window, reserving the final $T=120$ trading days for evaluation and using earlier data for training. The core component of the POMDP is an unobserved latent state defined per component as a \textbf{debit-only loss budget} (\textbf{health}), $s_t \in [0,100]$. Each component receives a small loss budget: on a day with a negative return, $s_t$ is debited proportionally and decreases; on a non-negative day, $s_t$ is unchanged; the state does not self-recover. Health increases only when the agent executes \textbf{recapitalize}. All actions draw from one shared, limited budget, and actions are taken when drift relative to the per-component no-loss floor becomes meaningful. This design is practice-inspired for two reasons. First, because we manage a large number of components, governance and our own policy favor a conservative stance: we avoid repeatedly allocating budget to components with recent serial losses, so the health is debit-only and does not auto-replenish. Second, it follows the risk-budgeting workflow described in \citet{meketa2024riskbudgeting}—set a budget ex ante, allocate and monitor against a benchmark, and treat material drift as a trigger for action. To make the benchmark operational, we instantiate a per-component \textbf{no-loss floor}: losses are deviations that consume the per-component budget; gains are consistent with the floor and do not raise limits by themselves; replenishment occurs only through \textbf{recapitalize}. For training and evaluation, components are assumed independent.

\noindent
\textbf{Actions and Costs:} The agent's action space $\mathcal{A}=\{\text{defer, inspect, recapitalize}\}$ manages the per-component loss budget (health). All actions draw from a shared, limited budget $B$ and follow a strict cost hierarchy $c_{\mathrm{recapitalize}}>c_{\mathrm{inspect}}>c_{\mathrm{defer}}$:
\begin{itemize}[itemsep=2pt, topsep=4pt, parsep=2pt]
    \item \textbf{Defer:} Continue with the current position. Incurs a low, continuous cost $c_{\mathrm{defer}}$ each step. Health remains subject to depletion by negative returns.
    \item \textbf{Inspect:} Pay $c_{\mathrm{inspect}}$ to obtain a precise observation of the hidden health $s_t$ for the selected component.
    \item \textbf{Recapitalize:} Pay the high cost $c_{\mathrm{recapitalize}}$ to rebuild health by resetting $s_t$ to $100$. This is the only action that increases health.
\end{itemize}

\noindent
\textbf{Objective and Failure Condition.} The agent's objective is to learn a policy $\pi$ that maximizes its \textbf{survival time}. An absorbing failure state is triggered immediately if any component’s health is exhausted, i.e., $s_t \le 0$. For each day the agent survives, it receives a reward of +1. This setup forces the agent to learn a sophisticated policy that balances the continuous drain from defer costs and market losses against the high, discrete costs of inspection and recapitalization, in order to prolong its survival.

\subsubsection{Analysis of Recapitalization Policy}
\noindent
We evaluate our approach on a \textbf{stock-level loss-budget} management task constructed from the S\&P~500 universe.
Starting from $500$ constituents, we retain the subset with at least $80\%$ daily-price coverage over the preceding three years, yielding $471$ components.
As in the infrastructure experiment, we reserve the final $T=120$ trading days for evaluation and use earlier data for model training.

\paragraph{State, actions, and costs.}
Each component $j$ is modeled as a single-component monotonic POMDP with an unobserved, debit-only \textbf{loss-budget (health)} $s_t^j \in [0,100]$.
Negative returns debit $s_t^j$ proportionally; non-negative returns leave $s_t^j$ unchanged; the state does not self-recover.
The action set is $\mathcal{A}=\{\text{defer},\text{inspect},\text{recapitalize}\}$ with a strict cost hierarchy $c_{\mathrm{recapitalize}}>c_{\mathrm{inspect}}>c_{\mathrm{defer}}$.
A global budget $B_{\text{tot}}$ is shared across all components. Table ~\ref{tab:etf_costs} presents the values of the various parameters used for the experiments.

\begin{table}[!htb]
\centering
\begin{tabular}{l c}
\toprule
Parameter & Value \\
\midrule
Total budget $B_{\text{tot}}$ & $15{,}000$ \\
Recapitalization cost $c_{\textsc{recap}}$ & $10.0$ \\
Inspection cost $c_{\textsc{insp}}$ & $0.5$ \\
Defer cost $c_{\textsc{def}}$ & $0.2$ \\
Number of components & $471$ \\
Evaluation horizon $T$ & $120$ days \\
\bottomrule
\end{tabular}
\caption{Cost and budget settings for the stock-level scenario.}
\label{tab:etf_costs}
\end{table}

\paragraph{Policies compared.}
We compare four policies:
\begin{enumerate}
    \item \textbf{Oracle}: full observability of the health $s_t$; \textbf{recapitalize} whenever $s_t<20$ (no inspection cost).
    \item \textbf{Oracle-guided meta-PPO}: the agent chooses \textbf{inspect} vs.\ \textbf{defer}; upon \textbf{defer}, it executes the oracle’s suggested restorative/default control; upon \textbf{inspect}, it pays $c_{\mathrm{inspect}}$ to reduce belief uncertainty. The agent learns when to buy observations and when to accept uncertainty.
    \item \textbf{Baseline (Heuristic)}: fixed \textbf{inspect} every $5$ trading days; if the observed $s_t<20$, take \textbf{recapitalize}; otherwise \textbf{defer}.
    \item \textbf{Vanilla meta-PPO}: trained on the same component–budget pairs as the oracle-guided agent but without oracle shaping.
\end{enumerate}

\paragraph{Budget allocation.}
We allocate $B_{\text{tot}}$ across the $471$ \textbf{components} using the same random-forest surrogate procedure as in the maintenance experiment: for each component $i$ we fit a concave surrogate for the map $B \mapsto \mathbb{E}[T^{i}_{\max}(B)]$ and solve a tractable concave maximization to obtain per-component budgets.

\paragraph{Training details.}
Both \textbf{oracle-guided meta-PPO} and \textbf{vanilla meta-PPO} are trained for $2\times10^{6}$ timesteps with Adam step size $10^{-4}$, minibatch size $128$, PPO horizon $T_{\mathrm{PPO}}=2048$, and discount factor $0.95$. For each component and policy we run $100$ simulations and report the component-level average $T^{i}_{\max}$; we then average across all $471$ components to obtain $\hat{T}_{\max}$.
\begin{figure}[!htb]
    \centering
    \includegraphics[width=0.7\textwidth]{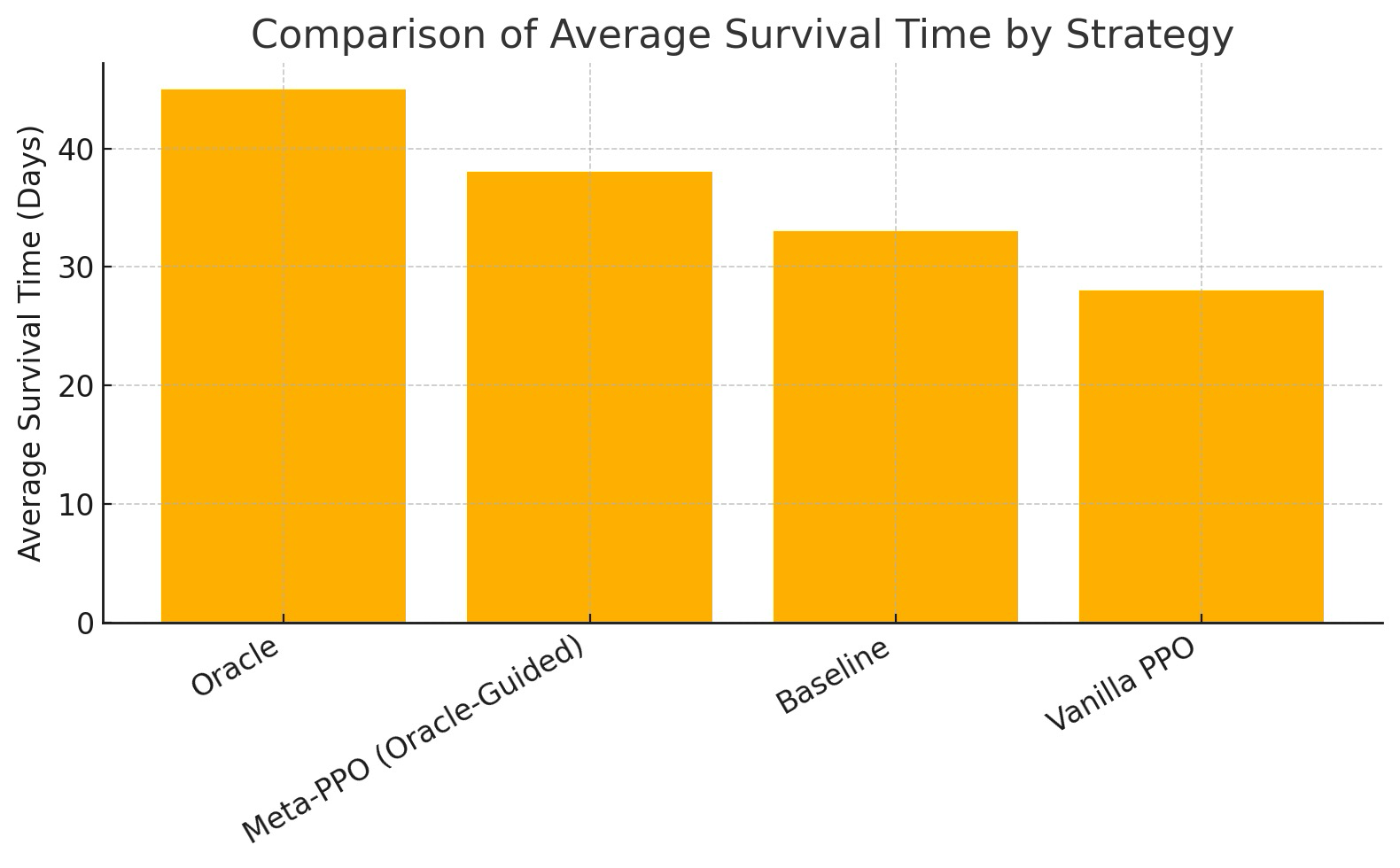 }
    \caption{S\&P 500 stock-level scenario: average survival time $\hat{T}_{\max}$ under a shared budget $B_{\mathrm{tot}}=15{,}000$ across 471 components. Observed ordering: \textbf{Oracle} $>$ \textbf{Oracle-guided meta-PPO} $>$ \textbf{Baseline} $>$ \textbf{Vanilla meta-PPO}.}
    \label{fig:etf_compare_metrics}
\end{figure}

\paragraph{Findings.}
Under the shared budget $B_{\text{tot}}=15{,}000$ across $471$ components and a $120$-day evaluation horizon, we observe a consistent ordering (see Figure~\ref{fig:etf_compare_metrics}):
\(\textbf{Oracle} > \textbf{Oracle-guided meta-PPO} > \textbf{Baseline} > \textbf{Vanilla meta-PPO}\).
The \textbf{Vanilla meta-PPO} tends to conserve budget and rarely recapitalizes, yielding the lowest survival time. The \textbf{Baseline} performs periodic inspections (every $5$ trading days) and recapitalizes below the threshold but spends budget indiscriminately and misses urgent cases. By contrast, the \textbf{Oracle-guided meta-PPO} learns when to inspect versus defer and when to act, allocating budget to higher-value opportunities; it reliably outperforms the Baseline and closes a substantial portion of the gap to the Oracle upper bound.

\subsubsection{Generalizability and Window Robustness of the Oracle-guided Meta-PPO}
\label{subsec:gen-robust-meta}

\paragraph{Design.}
We vary the number of \textbf{components} $N\in\{5,10,20,100,471\}$. For each $N$, the policy is trained on rolling $120$-day train windows and evaluated both in-sample (train) and on a held-out test window. We report average survival time (days) over $r{=}5$ seeds; error bars denote $\pm1$ standard deviation across seeds.

\begin{figure}[t]
  \centering
  \includegraphics[width=0.8\linewidth]{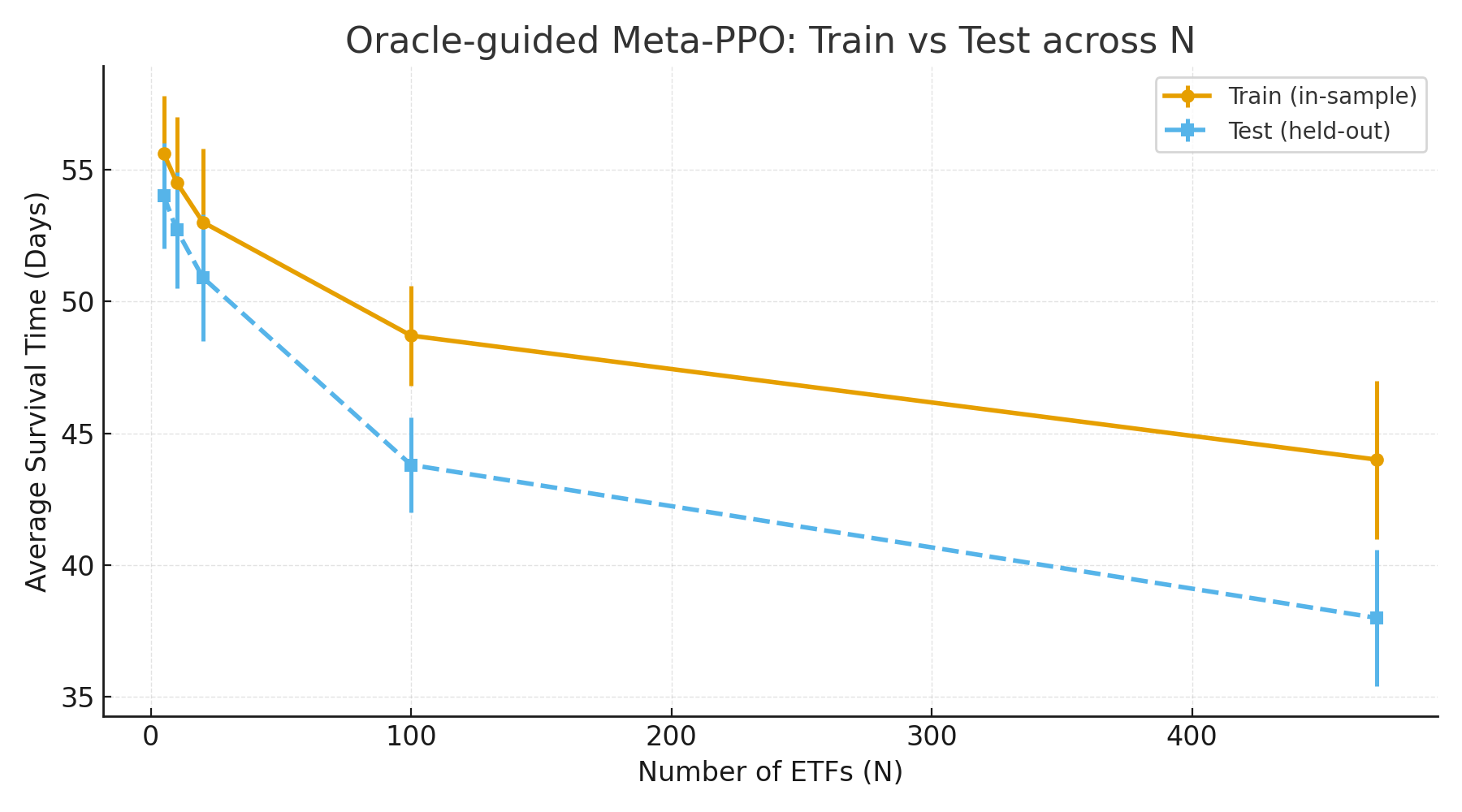}
  \caption{Oracle-guided meta-PPO: train vs.\ test across component set size $N$. The $y$-axis is average survival time (days); the $x$-axis is the number of components.}
  \label{fig:meta-ppo-train-test}
\end{figure}

\paragraph{Findings.}
(1) Both train and test curves decrease as $N$ grows, reflecting budget dilution and increased problem complexity. (2) Train performance is consistently above test with a modest generalization gap that tends to widen at larger $N$. (3) Variability is non-negligible and generally larger at higher $N$.

\paragraph{Takeaway.}
As can be seen from Figure~\ref{fig:meta-ppo-train-test}, the oracle-guided meta-PPO exhibits window robustness: trends are consistent across train windows, and the train-to-test drop remains moderate. At small $N$, the effective exploration/interaction budget is limited, which can hinder learning; as $N$ increases, richer allocation opportunities make better use of the oracle guidance even though absolute survival time declines under a fixed total budget.

\section{Conclusions}\label{sec:conclusion}
We proposed a scalable framework for solving \emph{budget-constrained
multi-component monotonic POMDPs}.  
Our chief theoretical contribution is a proof that the single-component
value function is \textbf{concave in budget}, which underpins an efficient
two-step solution strategy.  
First, a random-forest surrogate exploits that concavity to distribute the
shared budget across components, thereby decomposing the large $n$-component POMDP into $n$ independent single-component POMDPs. Second, an \emph{oracle-guided, meta-trained PPO} agent---shaped by value
iteration on the fully observable counterpart---learns a near-optimal policy
for each component–budget pair. Comprehensive experiments on two disparate domains confirm the framework’s
generality.  
For a 1000-component building-maintenance task, our method
significantly prolongs component survival relative to baseline heuristics
and approaches the performance of the oracle policy.  
On an ETF portfolio-rebalancing problem with draw-down–risk budgets, the
same algorithm consistently preserves portfolio viability and outperforms
vanilla PPO and the equal-weight baseline.  
Across both settings, empirical runtimes grow \emph{linearly} with the
number of components, validating the scalability predicted by our
complexity analysis. Future work will focus on extending the framework's capabilities to more dynamic budget allocation schemes and more complicated hierarchical budget constraints.

% \paragraph{Future work.}
% We plan to extend the framework to (i) dynamic or non-stationary budget
% allocation,\ (ii) hierarchical or multi-tier budget structures, and (iii)
% non-monotonic domains where components may partially recover without
% intervention.  Another promising direction is to replace random forests
% with differentiable surrogates, enabling end-to-end optimization of budget
% splits and control policies.

\bibliography{tmlr}
\bibliographystyle{tmlr}
\appendix
\section{Function Approximation of \texorpdfstring{$\mathbb{E}[T_{\max}]$}{E[Tmax]}}\label{appdx:fun_approx}
We model $\mathbb{E}[T^i_{\max}]$ as an exponential function of the budget allocated to component $i$. The choice of an exponential function is motivated by its ability to capture the saturation in $\mathbb{E}[T^i_{\max}]$ values at higher budget levels, a result of the finite planning horizon $H$. Additionally, the exponential model accounts for non-zero $\mathbb{E}[T^i_{\max}]$ even when the budget is zero. 
\begin{figure}[!ht]
    \centering
    \includegraphics[width=\linewidth]{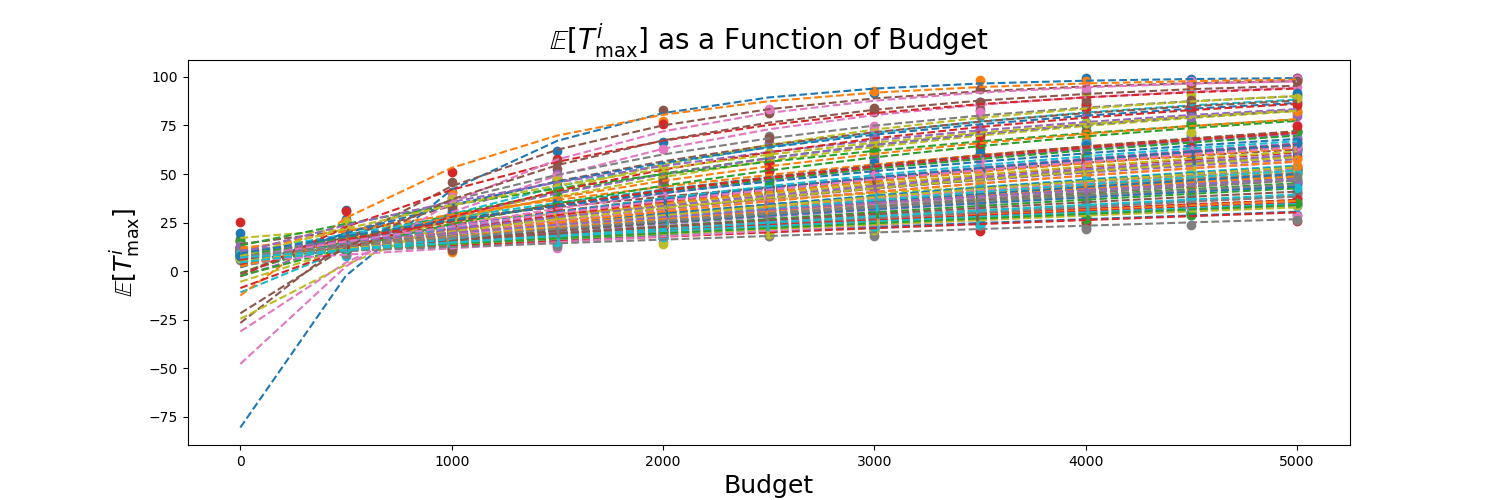}
    \caption{Exponential $\Tilde{T}^i_{\max}$ curves obtained using non-linear least-squares regression.}
    \label{fig:fig1}
\end{figure}

To validate the accuracy of this exponential model for $\mathbb{E}[T^i_{\max}]$, we conducted non-linear least squares regression on 100 infrastructure components. Figure~\ref{fig:fig1} illustrates the curves obtained through this regression, where $\mathbb{E}[T^i_{\max}]$ is modeled as an exponential function. The results indicate that the exponential function provides a strong approximation for $\mathbb{E}[T^i_{\max}]$, with an average coefficient of determination $R^2_{mean} = 0.899$.

\end{document}